\documentclass[conference,letterpaper,10pt]{IEEEtran}
\usepackage{fancyhdr,graphicx,times,amsmath,amssymb,cite,subfigure,color,url}
\usepackage[lined,ruled,commentsnumbered]{algorithm2e}
 \allowdisplaybreaks

\fancypagestyle{plain}{
            \fancyhf{}         % clear all header and footer fields
            \fancyfoot[L]{}
            \fancyfoot[C]{}
            \fancyfoot[R]{}

}

{
            \fancyhf{}
            \fancyfoot[R]{}}

\begin{document}

\title{Agreement Rate Initialized Maximum Likelihood Estimator for Ensemble Classifier Aggregation and Its Application in Brain-Computer Interface}

\author{\IEEEauthorblockN{Dongrui Wu\IEEEauthorrefmark{1},  Vernon J. Lawhern\IEEEauthorrefmark{2}\IEEEauthorrefmark{3}, Stephen Gordon\IEEEauthorrefmark{4}, Brent J. Lance\IEEEauthorrefmark{2},  Chin-Teng Lin\IEEEauthorrefmark{5}\IEEEauthorrefmark{6}}
\IEEEauthorblockA{\IEEEauthorrefmark{1}DataNova, NY USA}
\IEEEauthorblockA{\IEEEauthorrefmark{2}Human Research and Engineering Directorate, U.S. Army Research Laboratory, Aberdeen Proving Ground, MD USA}
\IEEEauthorblockA{\IEEEauthorrefmark{3}Department of Computer Science, University of Texas at San Antonio, San Antonio, TX USA}
\IEEEauthorblockA{\IEEEauthorrefmark{4}DCS Corp, Alexandria, VA USA}
\IEEEauthorblockA{\IEEEauthorrefmark{5}Brain Research Center, National Chiao-Tung University, Hsinchu, Taiwan}
\IEEEauthorblockA{\IEEEauthorrefmark{6}Faculty of Engineering and Information Technology, University of Technology, Sydney, Australia}
E-mail: drwu09@gmail.com, vernon.j.lawhern.civ@mail.mil, sgordon@dcscorp.com,\\ brent.j.lance.civ@mail.mil, ctlin@mail.nctu.edu.tw}
\IEEEoverridecommandlockouts
\IEEEpubid{\makebox[\columnwidth]{\hfill 978-1-5090-1897-0/16/\$31.00~\copyright2016 IEEE
                        } \hspace{\columnsep}\makebox[\columnwidth]{ }}
\maketitle

\begin{abstract}
Ensemble learning is a powerful approach to construct a strong learner from multiple base learners. The most popular way to aggregate an ensemble of classifiers is majority voting, which assigns a sample to the class that most base classifiers vote for. However, improved performance can be obtained by assigning weights to the base classifiers according to their accuracy. This paper proposes an agreement rate initialized maximum likelihood estimator (ARIMLE) to optimally fuse the base classifiers. ARIMLE first uses a simplified agreement rate method to estimate the classification accuracy of each base classifier from the unlabeled samples, then employs the accuracies to initialize a maximum likelihood estimator (MLE), and finally uses the expectation-maximization algorithm to refine the MLE. Extensive experiments on visually evoked potential classification in a brain-computer interface application show that ARIMLE outperforms majority voting, and also achieves better or comparable performance with several other state-of-the-art classifier combination approaches.
\end{abstract}


\section{Introduction}

Ensemble learning \cite{Dietterich2000, Breiman1998, Hashem1997} is very effective in constructing a strong learner from multiple base (weak) learners, for both classification and  regression problems. This paper focuses on ensemble learning for binary classification problems. More specifically, we investigate how to optimally combine multiple base binary classifiers for better performance.

Given an ensemble of base binary classifiers, the simplest yet most popular ensemble learning approach is majority voting (MV), i.e., assigning a sample to the class that most base classifiers agree on. However, the base classifiers usually have different classification accuracies, and hence considering them equally (as in MV) in aggregation may not be optimal. It is more intuitive to use weighted voting, where the weight is a function of the corresponding classification accuracy.

The first step in weighted voting is to estimate the accuracies of the base classifiers. There could be two approaches. The first is to use cross-validation on the training data. However, in many applications the training data may be very limited, so the cross-validation accuracy may not be reliable. For example, in the brain-computer interface (BCI) system calibration application considered in this paper (Section~\ref{sect:experiments}), to increase the utility of the BCI system, we would like to use as little calibration data as possible, preferably zero; so, it is difficult to perform cross-validation. Moreover, in certain situations only the output of the classifiers are available. Thus, it is not feasible to perform cross-validation.

Because of these limitations, in this paper we consider the second approach, in which the accuracies of the base classifiers are estimated from their predictions on the unlabeled samples. There have been a few studies in this direction. Platanios et al. \cite{Platanios2014} used agreement rate (AR) among different base classifiers to estimate both the marginal and joint error rates (However, they did not show how the error rates can be used to optimally combine the classifiers). Parisi et al. \cite{Parisi2014} proposed a spectral meta-learner (SML) approach to estimate the accuracies of the base classifiers from their population covariance matrix, and then used them in a maximum likelihood estimator (MLE) to aggregate these base classifiers. Researchers from the same group then proposed several different approaches \cite{Jaffe2015,Jaffe2015b,Shaham2016} to improve the SML. They have all shown better performance than MV.

This paper proposes a new classifier combination approach, agreement rate initialized maximum likelihood estimator (ARIMLE), to aggregate the base classifiers. As its name suggests, it first uses the AR method to estimate the classifier accuracies, and then employs them in an MLE to optimally fuse the classifiers. Using a visually evoked potential (VEP) BCI experiment with 14 subjects and three different EEG headsets, we show that ARIMLE outperforms MV, and its performance is also better than or comparable to several other state-of-the-art classifier combination approaches.

The remainder of the paper is organized as follows: Section~\ref{sect:ARIMLE} introduces the details of the ARIMLE algorithm. Section~\ref{sect:experiments} describes experiment setup and performance comparisons of eight different algorithms. Section~\ref{sect:conclusions} draws conclusions.

\section{ARIMLE for Classifier Aggregation} \label{sect:ARIMLE}

This section introduces the proposed ARIMLE for classifier aggregation.

\subsection{Problem Setup}

The problem setup is very similar to that in \cite{Parisi2014,Jaffe2015,Jaffe2015b,Shaham2016}, so we use similar notations and terminology.

We consider binary classification problems with input space $\mathcal{X}$ and output space $\mathcal{Y}\in\{-1,1\}$. A sample and class label pair $(X,Y)\in\mathcal{X}\times\mathcal{Y}$ is a random vector with joint probability density function $p(\mathbf{x},y)$, and
marginal probability density functions $p_X(\mathbf{x})$ and $p_Y(y)$. Assume there are $n$ unlabeled samples, $\{\mathbf{x}_j\}_{j=1}^n$, with unknown true labels $\{y_j\}_{j=1}^n$. Assume also there are $m$ base binary classifiers, $\{f_i\}_{i=1}^m$, and the $i$th classifier's prediction for $\mathbf{x}_j$ is $f_i(\mathbf{x}_j)$. Define the classification sensitivity of $f_i$ as
\begin{align}
\psi_i=\mathrm{P}(f_i(X)=1|Y=1) \label{eq:psi}
\end{align}
and its specificity as
\begin{align}
\eta_i=\mathrm{P}(f_i(X)=-1|Y=-1) \label{eq:eta}
\end{align}
Then, the balanced classification accuracy (BCA) of $f_i$ is
\begin{align}
\pi_i=\frac{1}{2}(\psi_i+\eta_i).
\end{align}

As in \cite{Parisi2014}, we make two important assumptions in the following derivation: 1) The $n$ unlabeled samples $\{\mathbf{x}_j\}_{j=1}^n$ are independent and identically distributed realizations from $p_X(\mathbf{x})$; and, 2) The $m$ base binary classifiers $\{f_i\}_{i=1}^m$ are independent, i.e., prediction errors made by one classifier are independent of those made by any other classifier.

\subsection{Agreement Rate (AR) Computation}

The AR method presented in this subsection is a simplified version of the one introduced in \cite{Platanios2014}, by assuming any pair of $f_{i_1}$ and $f_{i_2}$ ($i_1\neq i_2$) are independent. It is used to compute the (unbalanced) error rate of each classifier, which is defined as
\begin{align}
e_i=\mathrm{P}(f_i(X)\neq Y), \quad i=1,...,m
\end{align}
which is in turn used in the next subsection to construct the MLE.

We define the AR of two classifiers $f_{i_1}$ and $f_{i_2}$ $(i_1\neq i_2$) as the probability that they give identical outputs, i.e.,
\begin{align}
a_{i_1,i_2}=\mathrm{P}(f_{i_1}(X)=f_{i_2}(X)) \label{eq:a0}
\end{align}
which can be empirically computed from the predictions of the two classifiers.

As in \cite{Platanios2014}, we can show that
\begin{align}
a_{i_1,i_2}=1-e_{i_1}-e_{i_2}+2e_{i_1,i_2} \label{eq:a1}
\end{align}
where $e_{i_1,i_2}$ is the (unbalanced) joint error rate of $f_{i_1}$ and $f_{i_2}$. Under the assumption that $f_{i_1}$ and $f_{i_2}$ are independent, we have $e_{i_1,i_2}=e_{i_1}\cdot e_{i_2}$, and hence (\ref{eq:a1}) can be re-expressed as:
\begin{align}
a_{i_1,i_2}=1-e_{i_1}-e_{i_2}+2e_{i_1}\cdot e_{i_2} \label{eq:a2}
\end{align}

To find the $m$ error rates for the $m$ classifiers, we compute the AR $a_{i_1,i_2}$ for all $\frac{1}{2}m(m-1)$ possible combinations of $(i_1,i_2)$, $i_1=1,...,m$, $i_2=1,...,m$, and $i_1\neq i_2$. By substituting them into (\ref{eq:a2}), we have $\frac{1}{2}m(m-1)$ equations and $m$ variables $\{e_i\}_{i=1}^m\in[0,1]$, which can be easily solved by a constrained optimization routine, e.g., \emph{fmincon} in Matlab.

The main difference between our approach for estimating $\{e_i\}_{i=1}^m$ and the one in \cite{Platanios2014} is that, \cite{Platanios2014} considers the general case that different base classifiers are inter-dependent, and hence it tries to find $2^m-1$ error rates ($m$ marginal error rates $\{e_i\}_{i=1}^m$ for the individual classifiers, $\frac{1}{2}m(m-1)$ joint error rates $\{e_{i_1,i_2}\}_{i_1\neq i_2}$ for all pairs of classifiers, $\frac{1}{6}m(m-1)(m-2)$ joint error rates $\{e_{i_1,i_2,i_3}\}_{i_1\neq i_2\neq i_3}$ for all 3-tuples of classifiers, and so on) all at once. Since there are more error rates than equations, it introduces additional constraints, e.g., to minimize the dependence between different classifiers, to solve for the $2^m-1$ error rates. We do not adopt that approach because of its high computational cost. For example, in our experiments in Section~\ref{sect:experiments} we have 13 base classifiers, i.e., $m=13$, and hence $2^m-1=8191$ error rates to optimize, which is very computationally expensive. So, we make the simplified assumption that all $m$ base classifiers are independent, and hence only need to find the $m$ marginal error rates $\{e_i\}_{i=1}^m$. $\{e_i\}_{i=1}^m$ estimated here may not be as accurate as the ones in \cite{Platanios2014}, but they are only used to initialize our MLE, and in the next subsection we shall use an expectation-maximization (EM) algorithm to iteratively improve them.

Once $\{e_i\}_{i=1}^m$ are obtained, the (unbalanced) classification accuracy of $f_i$ is then computed as $1-e_i$, which is also an estimate of the BCA $\pi_i$, i.e.,
\begin{align}
\pi_i\approx 1-e_i, \quad i=1,...,m \label{eq:pi}
\end{align}
by assuming that the positive and negative classes have similar accuracies.

\subsection{Maximum Likelihood Estimator (MLE)}

As shown in \cite{Parisi2014}, the MLE from $\{f_i\}_{i=1}^m$ is
\begin{align}
\hat{y}=\mathrm{sign}\left[\sum_{i=1}^m \left(f_i(\mathbf{x})\ln\alpha_i+\ln\beta_i\right)\right] \label{eq:yMLE}
\end{align}
where
\begin{align}
\alpha_i&=\frac{\psi_i\eta_i}{(1-\psi_i)(1-\eta_i)} \label{eq:alpha}\\
\beta_i&=\frac{\psi_i(1-\psi_i)}{\eta_i(1-\eta_i)} \label{eq:beta}
\end{align}
i.e., the MLE is a linear ensemble classifier, whose weights depend on the unknown specificities and sensitivities of the $m$ base classifiers.

The classical approach for solving (\ref{eq:yMLE}) is to jointly maximize the likelihood for all $\{\hat{y}_j\}_{j=1}^n$, $\{\psi_i\}_{i=1}^m$ and $\{\eta_i\}_{i=1}^m$ using an EM algorithm \cite{Dawid1979,Raykar2010,Parisi2014,Welinder2010,Sheng2008,Whitehill2009}, which first estimates $\{\psi_i\}_{i=1}^m$ and $\{\eta_i\}_{i=1}^m$ given some initial $\{\hat{y}_j\}_{j=1}^n$, and then updates $\{\hat{y}_j\}_{j=1}^n$ using the newly estimated $\{\psi_i\}_{i=1}^m$ and $\{\eta_i\}_{i=1}^m$, and iterates until they converge. The question is how to find a good initial estimate of $\{\hat{y}_j\}_{j=1}^n$ so that the final estimates are less likely to be trapped in a local minimum.

We solve this problem by using the results from \cite{Parisi2014}, which suggested that the BCAs $\{\pi_i\}_{i=1}^m$ can be used to compute a good initialization of $\{\hat{y}_j\}_{j=1}^n$, i.e.,
\begin{align}
\hat{y}_j=\mathrm{sign}\left[\frac{\sum_{i=1}^m (2\pi_i-1) f_i(\mathbf{x}_j)}{\sum_{i=1}^m(2\pi_i-1)}\right],\quad j=1,...,n \label{eq:yhat0}
\end{align}
The EM algorithm can then run from there.

\subsection{The Complete ARIMLE Algorithm}

The complete ARIMLE algorithm is shown in Algorithm~1. It first uses AR to compute the error rate of each base classifier, then employs the error rates to initialize the EM algorithm, and finally runs the EM algorithm until a stopping criterion is met, which could be reaching the maximum number of iterations, or the difference between the last two iterations is smaller than a certain threshold. The former is used in this paper.

\begin{algorithm}[h] %\DontPrintSemicolon
\KwIn{$n$ unlabeled samples, $\{\mathbf{x}_j\}_{j=1}^n$\;
\hspace*{10mm} $m$ base binary classifiers, $\{f_i\}_{i=1}^m$.}
\KwOut{The maximum likelihood estimates $\{\hat{y}\}_{j=1}^n$.}
\For{$i_1=1,...,m-1$}{
\For{$i_2=i_1+1,...,m$}{
Compute $a_{i_1,i_2}$ in (\ref{eq:a0})\; }
Solve for $\{e_i\}_{i=1}^m$ in (\ref{eq:a2}) using constrained optimization\;
Compute $\{\pi_i\}_{i=1}^m$ using (\ref{eq:pi})\;
}
Initialize $\{\hat{y}_j\}_{j=1}^n$ using (\ref{eq:yhat0})\;
\While{stopping criterion not met}{
Compute $\{\psi_i\}_{i=1}^m$ in (\ref{eq:psi}) and $\{\eta_i\}_{i=1}^m$ in (\ref{eq:eta}), by treating $\{\hat{y}_j\}_{j=1}^n$ as the true labels\;
Compute $\{\alpha_i\}_{i=1}^m$ in (\ref{eq:alpha}) and $\{\beta_i\}_{i=1}^m$ in (\ref{eq:beta})\;
Update $\{\hat{y}_j\}_{j=1}^n$ using (\ref{eq:yMLE})\; }
\textbf{Return} The latest $\{\hat{y}_j\}_{j=1}^n$.
\caption{The ARIMLE algorithm.} \label{alg:ARIMLE}
\end{algorithm}

\section{Experiments and Analysis} \label{sect:experiments}

This section presents the experiment setup that is used to evaluate the performance of ARIMLE, and the performance comparison of ARIMLE with MV and several other state-of-the-art classifier combination approaches.

\subsection{Experiment Setup}

\begin{figure}[htpb]\centering
\subfigure[]{\label{fig:T}     \includegraphics[width=.22\linewidth]{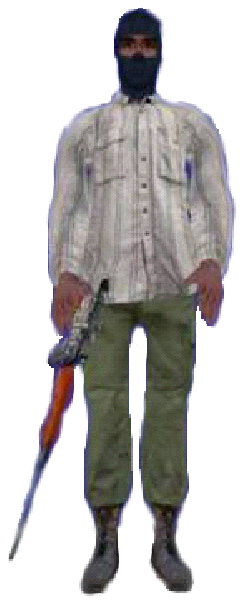}}
\subfigure[]{\label{fig:NT}     \includegraphics[width=.20\linewidth]{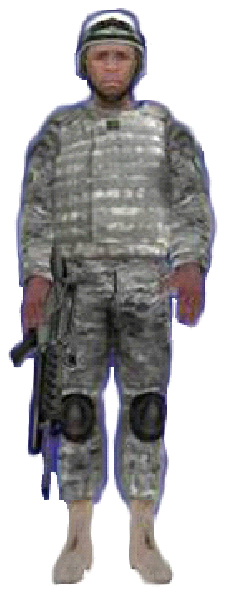}}
\caption{Example images of (a) a target; (b) a non-target.} \label{fig:TNT}
\end{figure}

We used data from a VEP oddball task \cite{Ries2014}. Image stimuli of an enemy combatant [target, as shown in Fig.~\ref{fig:T}] or a U.S. Soldier [non-target, as shown in Fig.~\ref{fig:NT}] were presented to subjects at a rate of 0.5 Hz. The subjects were instructed to identify each image as being target or non-target with a unique button press as quickly and accurately as possible. There were a total of 270 images, of which 34 were targets. The experiments were approved by the U.S. Army Research Laboratory (ARL) Institutional Review Board (Protocol \# 20098-10027). The voluntary, fully informed consent of the persons used in this research was obtained as required by federal and Army regulations \cite{USArmy,USDoD}. The investigator adhered to Army policies for the protection of human subjects.

Eighteen subjects participated in the experiments, which lasted on average 15 minutes. Data from four subjects were not used due to data corruption or lack of responses. Signals from each subject were recorded with three different EEG headsets, including a wired 64-channel 512Hz ActiveTwo system from BioSemi, a wireless 9-channel 256Hz B-Alert X10 EEG Headset System from Advanced Brain Monitoring (ABM), and a wireless 14-channel 128Hz EPOC headset from Emotiv.

\subsection{Preprocessing and Feature Extraction}

The EEG data preprocessing and feature extraction methods were similar to those used in \cite{drwuTNSRE2016,drwuSMC2015}. EEGLAB \cite{Delorme2004} were used to extract raw EEG amplitude features.

For each headset, we first band-passed the EEG signals to [1, 50] Hz, then downsampled them to 64 Hz, performed average reference, and next epoched them to the $[0, 0.7]$ second interval timelocked to stimulus onset. We removed mean baseline from each channel in each epoch and removed epochs with incorrect button press responses\footnote{Button press responses were not recorded for the ABM headset, so we used all epochs from it.}. The final numbers of epochs from the 14 subjects are shown in Table~\ref{tab:epoch}. Observe that there is significant class imbalance for all headsets.

\begin{table*}[htpb] \centering \setlength{\tabcolsep}{1mm}
\caption{Number of epochs for each subject after preprocessing. The numbers of target epochs are given in the parentheses.}   \label{tab:epoch}
\begin{tabular}{l|cccccccccccccc}   \hline
   Subject  &  1&2&3&4&5&6&7&8&9&10&11&12&13 &14 \\ \hline
   BioSemi &  241(26)&260(24)& 257(24) & 261(29)& 259(29)& 264(30)& 261(29) & 252(22)& 261(26)& 259(29)& 267(32)& 259(24)&261(25)& 269(33)\\
  Emotiv  &263(28) &  265(30) &  266(30)& 255(23)& 264(30)& 263(32)& 266(30)&252(22)& 261(26)& 266(29)& 266(32)& 264(33) & 261(26)& 267(31)\\
  ABM & 270(34) & 270(34) & 235(30) & 270(34) & 270(34)&270(34)&270(34)&270(33)&270(34)&239(30)&270(34)&270(34)&251(31)&270(34)\\
  \hline
\end{tabular}
\end{table*}

Each [0, 0.7] second epoch contains 45 raw EEG magnitude samples. The concatenated feature vector has hundreds of dimensions. To reduce the dimensionality, we performed a simple principal component analysis, and took only the scores for the first 20 principal components. We then normalized each feature dimension separately to $[0, 1]$ for each subject.

\subsection{Evaluation Process and Performance Measures} \label{sect:process}

\begin{figure}[htpb]\centering
 \includegraphics[width=.7\linewidth]{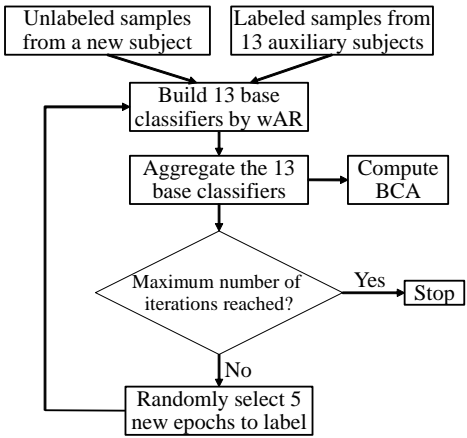} \caption{Flowchart of the evaluation process.} \label{fig:flowchart}
\end{figure}

Although we knew the labels of all EEG epochs from all headsets for each subject, we simulated a different scenario, as shown in Fig.~\ref{fig:flowchart}: None of the epochs from the current subject under study was initially labeled, but all epochs from all the other 13 subjects with the same headset were labeled. Our approach was to iteratively label some epochs from the current subject, and then to build an ensemble of 13 classifiers (one from each of the 13 auxiliary subjects) to label the rest of the epochs. Seven different algorithms (see the next subsection), including ARIMLE, were used to the aggregate the 13 classifiers. The goal was to achieve the highest BCA for the new subject, with as few labeled epochs as possible. Each classifier in the ensemble was constructed using the weighted adaptation regularization (wAR) algorithm in \cite{drwuSMC2015}, which is a domain adaptation approach in transfer learning.

In each iteration five epochs were labeled, and the algorithm terminated after 20 iterations, i.e., after 100 epochs were labeled. We repeated this process 30 times for each subject and each headset so that statistically meaningful results could be obtained.

The BCA was used as our performance measure.

\subsection{Algorithms}

We compare our propose ARIMLE with a baseline algorithm and several other state-of-the-art classifier combination approaches in the literature:
\begin{enumerate}
\item Baseline (BL), which uses only available labeled subject-specific data to train a support vector machine classifier and then applies it to the remaining unlabeled data.
\item MV, which computes the final label as $\hat{y}_j=\mathrm{sign}\left[\sum_{i=1}^mf_i(\mathbf{x}_j)\right]$, $j=1,...,n$. This is the most popular and also the simplest ensemble combination approach in the literature and practice.
\item Spectral meta-learner (SML) \cite{Parisi2014}, which estimates the BCAs of the base classifiers from their population covariance matrix, and then uses them in (\ref{eq:yhat0}) to compute the final estimates. There is no iterative EM algorithm involved.
\item Iterative MLE (iMLE) \cite{Parisi2014}, which performs the above SML first and then uses an EM algorithm to refine the MLE.
\item Improved SML (i-SML) \cite{Jaffe2015}, which first estimates the class imbalance of the labels and then uses that to directly estimate the sensitivity and specificity of each base classifier. The sensitivities and specificities are then used to construct the MLE.
\item Latent SML (L-SML) \cite{Jaffe2015b}, which, instead of assuming all $m$ classifiers are conditionally independent, assumes the $m$ classifiers can be partitioned into several groups according to a latent variable: the classifiers in the same group can be correlated, but the classifiers from different groups are conditionally independent. It is hoped that in this way it can better handle correlated base classifiers.
\end{enumerate}
Additionally, we also constructed an oracle SML (O-SML), which assumes that we know the true sensitivity and specificity of each base classifier, to represent the upper bound of the classification performance we could get from these $m$ base classifiers using MLE.

\subsection{Experimental Results and Discussions}

\begin{figure}[htpb]\centering
\subfigure[]{\label{fig:A}     \includegraphics[width=.92\linewidth,clip]{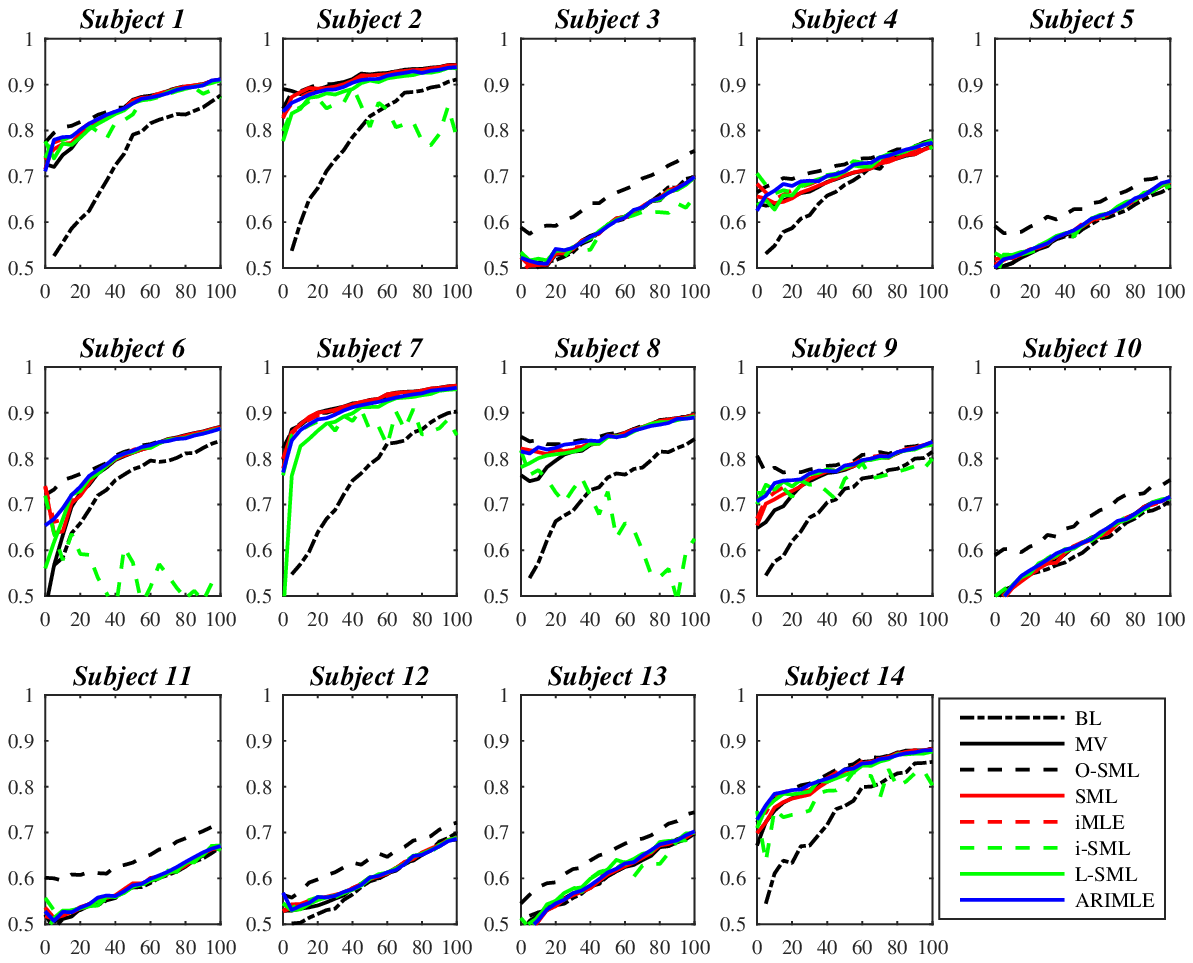}}
\subfigure[]{\label{fig:B}     \includegraphics[width=.92\linewidth,clip]{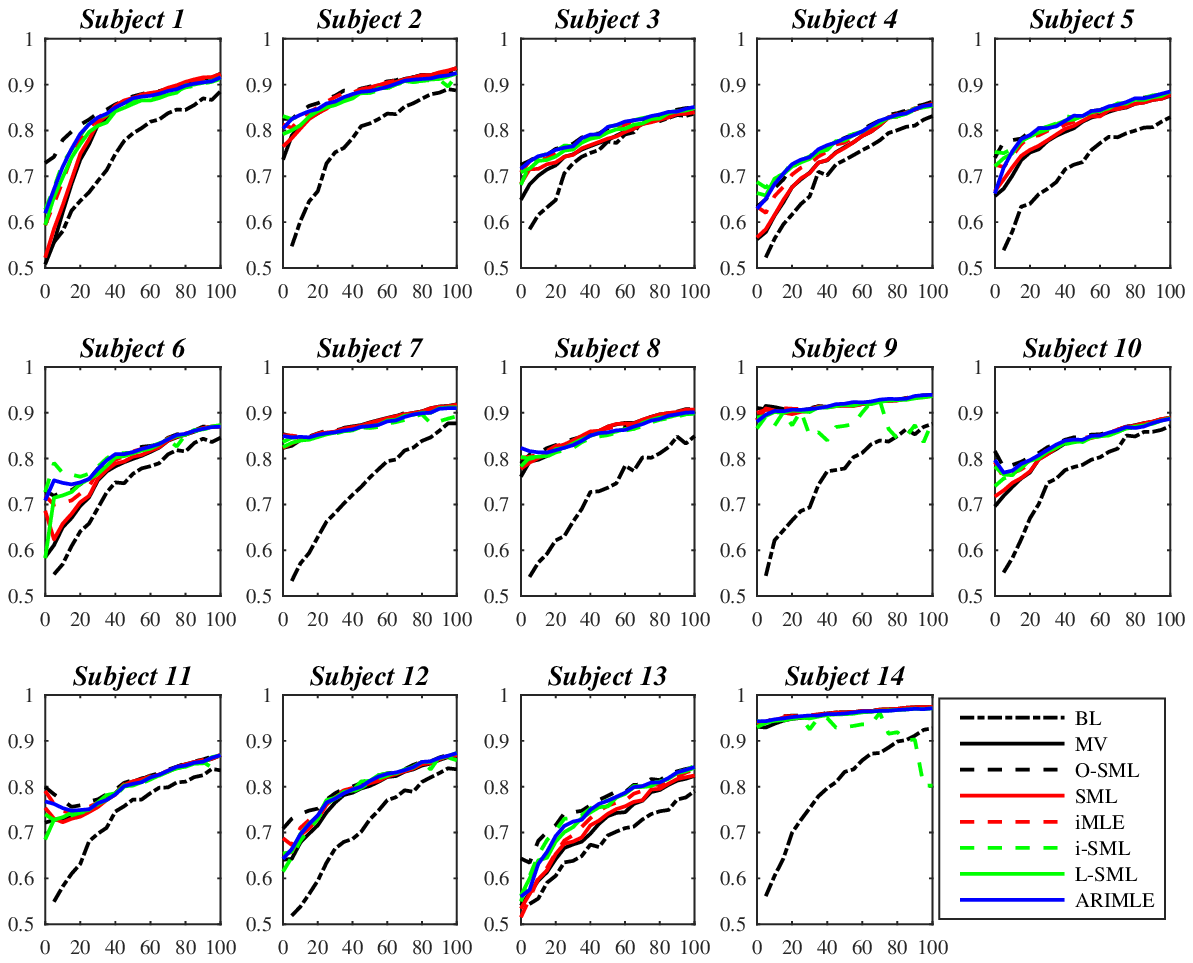}}
\subfigure[]{\label{fig:E}     \includegraphics[width=.92\linewidth,clip]{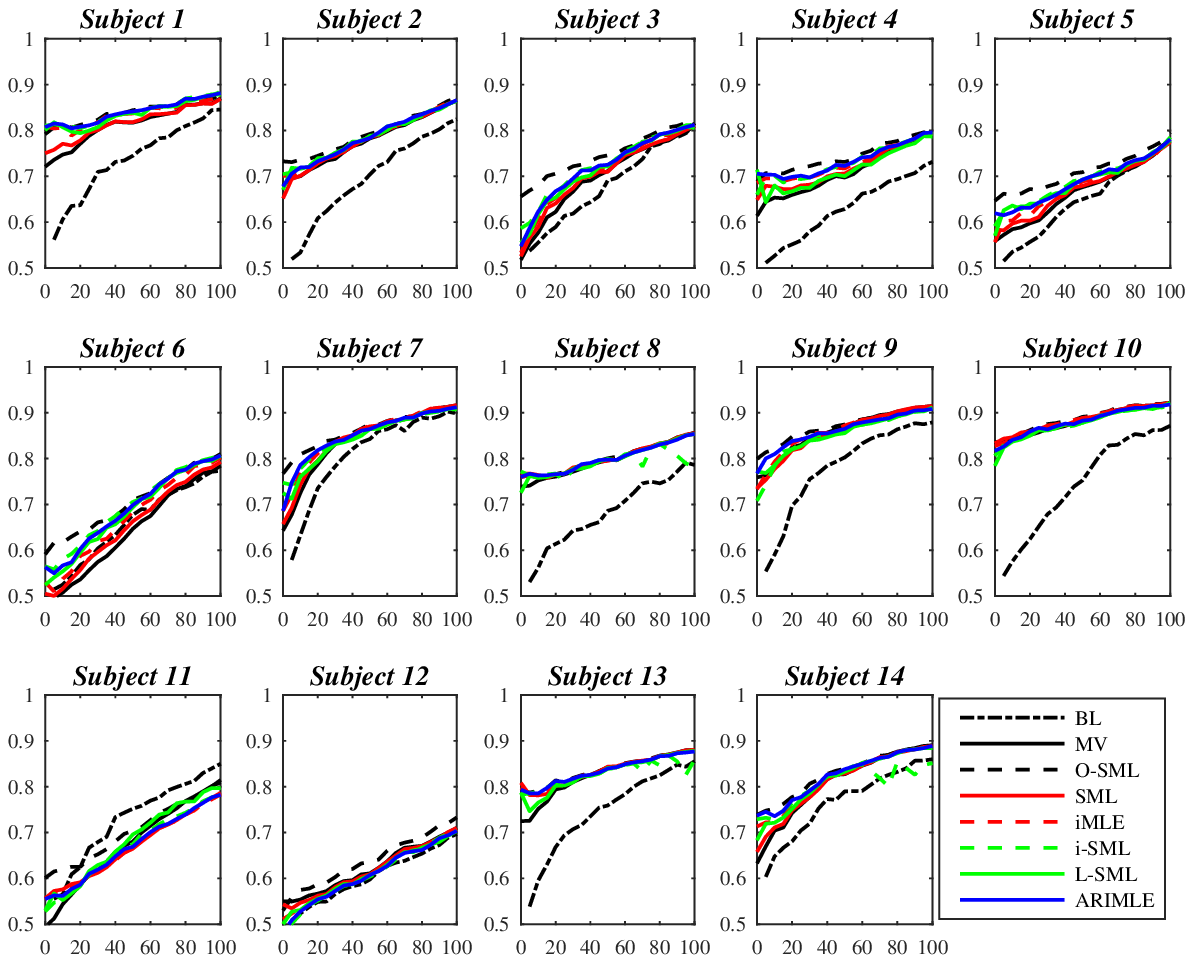}}
\caption{Individual BCAs of the eight algorithms for the 14 subjects, averaged over 30 runs for each headset. (a) ABM headset; (b) BioSemi headset; (c) Emotiv headset. Horizontal axis: $n_l$, the number of labeled epochs from the subject. Vertical axis: BCA.} \label{fig:ind}
\end{figure}

The average BCAs of the eight algorithms across the 14 subjects and three EEG headsets are shown in Fig.~\ref{fig:avgs}, along with the average performances across the three headsets, where $n_l$ denotes the number of labeled samples from the new subject. The accuracies for each individual subject, averaged over 30 runs, are shown in Fig.~\ref{fig:ind}. Non-parametric multiple comparison tests using Dunn's procedure \cite{Dunn1961,Dunn1964} were also performed on the combined data from all the subjects and headsets to determine if the difference between any pair of algorithms was statistically significant, with a $p$-value correction using the False Discovery Rate method by \cite{Benjamini1995}. The results are shown in Table~\ref{tab:DunnABE}, with the statistically significant ones marked in bold. Observe that:

\begin{figure}[htpb]\centering
\subfigure[]{\label{fig:Aavg}     \includegraphics[width=.48\linewidth,clip]{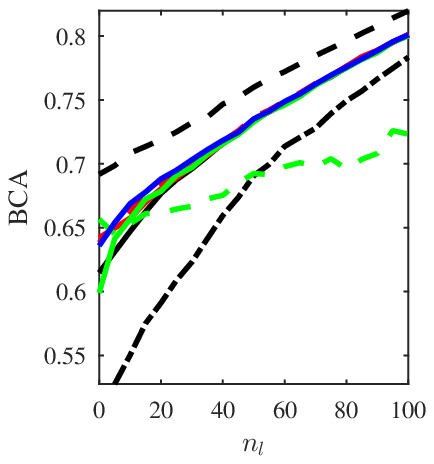}}
\subfigure[]{\label{fig:Bavg}     \includegraphics[width=.48\linewidth,clip]{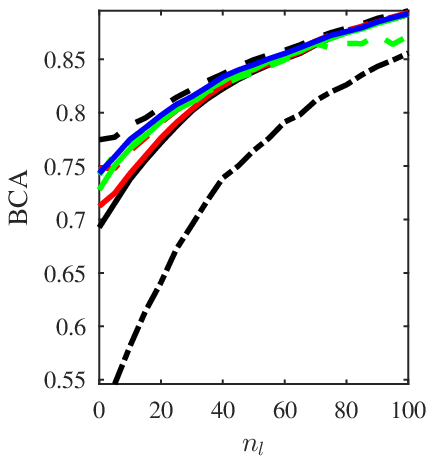}}
\subfigure[]{\label{fig:Eavg}     \includegraphics[width=.48\linewidth,clip]{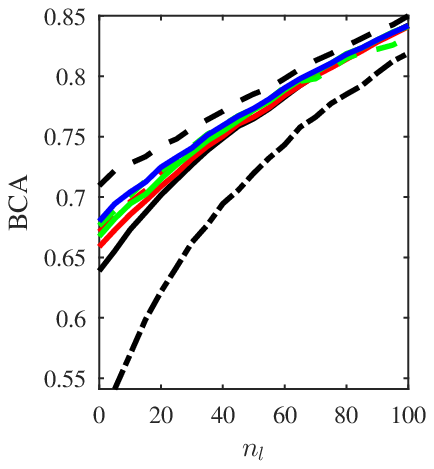}}
\subfigure[]{\label{fig:3avg}     \includegraphics[width=.48\linewidth,clip]{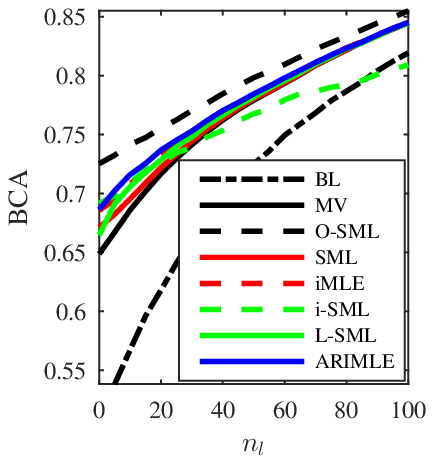}}
\caption{Average BCAs of the eight algorithms across the 14 subjects. (a) ABM headset; (b) BioSemi headset; (c) Emotiv headset; (d) average of the three headsets.} \label{fig:avgs}
\end{figure}

\begin{enumerate}
\item ARIMLE had significantly better performance than BL, which did not use transfer learning and ensemble learning. In fact, almost all seven algorithms based on transfer learning and ensemble learning achieved much better performance than BL.
\item ARIMLE almost always outperformed MV, SML and L-SML, and the performance improvement was statistically significant for small $n_l$.
\item ARIMLE had comparable performance with iMLE. For small $n_l$, the BCA of ARIMLE was slightly higher than iMLE. The performance difference was not statistically significant, but very close to the threshold.
\item i-SML gave good performance for most subjects, but sometimes the predictions were significantly off-target\footnote{We used our own implementation, and also Shaham et al.'s implementation \cite{Shaham2016} at https://github.com/ushaham/RBMpaper. The results were similar.}. Overall, ARIMLE outperformed i-SML.
\item O-SML outperformed ARIMLE, and the performance difference was statistically significant when $n_l$ is small, which suggests that there is still room for ARIMLE to improve: if the sensitivity and specificity of the base binary classifiers can be better estimated, then the performance of ARIMLE could further approach O-SML. This is one of our future research directions.
\end{enumerate}
In summary, we have shown through extensive experiments that ARIMLE significantly outperformed MV, and its performance was also better than or comparable to several state-of-the-art classifier combination approaches. Although a BCI application was considered in this paper, we believe the applicability of ARIMLE is far beyond that.

\begin{table}[htpb] \centering \setlength{\tabcolsep}{2mm}
\caption{$p$-values of non-parametric multiple comparisons of the BCA of ARIMLE versus other seven algorithms.}   \label{tab:DunnABE}
\begin{tabular}{c|ccccccc}   \hline
$n_l$   &            BL &             MV  &           O-SML &            SML & iMLE           &          i-SML & L-SML      \\ \hline
0  & N/A&  \textbf{.0000} & \textbf{.0000}  & \textbf{.0009} &          .4750 &    .0698 & \textbf{.0000}\\
5  & \textbf{.0000}&  \textbf{.0000} & \textbf{.0000}  & \textbf{.0000} &          .0466 &.4565 & .0266\\
10  & \textbf{.0000}& \textbf{.0000}  & \textbf{.0000}  &  \textbf{.0000} &          .0546 & .2410 & \textbf{.0191}\\
15  & \textbf{.0000}&   \textbf{.0000}& \textbf{.0001}  & \textbf{.0001} &          .0832 & .3063 & .0550\\
20  & \textbf{.0000}&  \textbf{.0000} &  \textbf{.0005} & \textbf{.0010} & .1683          &  .1470& .0460\\
25  & \textbf{.0000}&  \textbf{.0011} &  \textbf{.0015} & \textbf{.0083} & .2398          &  .1684& .1306\\
30  & \textbf{.0000}&  \textbf{.0202}  & \textbf{.0014}  & .0645          & .4245          &  .1412& .1735\\
35  & \textbf{.0000}& .0428           &  \textbf{.0080} & .1026          & .3781          &  .1436& .2437\\
40  & \textbf{.0000}& .0801           &   \textbf{.0163}& .1228          & .3734          &   .0847& .2150\\
45  & \textbf{.0000}& .1546           &  \textbf{.0117} & .2214          & .4656          & .1121 & .3344\\
50 &\textbf{.0000}& .2126           &  \textbf{.0105} & .2581          & .4386          &  .0370& .2503\\
55 &\textbf{.0000} & .2340           &  \textbf{.0199}          & .2528          & .4816          &  .0352 & .3380\\
60 &\textbf{.0000} &     .2707       &  .0359 &    .2972      &    .4650       &  .0291& .3073\\
65 &\textbf{.0000} &     .3110       &  .0331 &    .3107       &    .4908       &  .0306& .3201\\
70 &\textbf{.0000} &     .4088       &  .0263 &    .4222       &    .4682       &  \textbf{.0091}& .4287\\
75 &\textbf{.0000} &     .4815       &  .0403 &    .4418       &     .4777      &  \textbf{.0028}&.4046\\
80 &\textbf{.0000} &    .5060        &  .0355          &    .5442       &   .4582        &  \textbf{.0008}& .5331\\
85 & \textbf{.0000}&   .4985         &  .0336 &    .4813       &     .4857      &  \textbf{.0004}&.4733\\
90 &\textbf{.0000} &      .4706      & .0434           &   .4885        &   .4522        &  \textbf{.0002}&.4625\\
95 & \textbf{.0000}&    .5057        &  .0690 &   .4978        &   .5165        &  \textbf{.0001}& .5367\\
100 & \textbf{.0000}&    .4674        &  .0436     &   .4842        &   .4792        &  \textbf{.0000}& .4890\\   \hline
\end{tabular}
\end{table}

\section{Conclusions} \label{sect:conclusions}

This paper has proposed an ARIMLE approach to optimally aggregate multiple base binary classifiers in ensemble learning. It first uses AR to estimate the classification accuracies of the base classifiers from the unlabeled samples, which are then used to initialize an MLE. An EM algorithm is then employed to refine the MLE. Extensive experiments on visually evoked potential classification in a BCI application, which involved 14 subjects and three different EEG headsets, showed that ARIMLE significantly outperformed MV, and its performance was also better than or comparable to several other state-of-the-art classifier combination approaches. We expect ARIMLE to have broad applications beyond BCI.

Our future research will investigate the integration of ARIMLE with other machine learning approaches for more performance improvement. We have shown in \cite{drwuSMC2014,drwuTNSRE2016} that active learning \cite{Settles2009} can be combined with transfer learning to improve the offline classification performance: active learning optimally selects the most informative unlabeled samples to label (rather than random sampling), and transfer learning combines subject-specific samples with labeled samples from similar/relevant tasks to build better base classifiers. ARIMLE is an optimal classifier combination approach, which is independent of and also complementary to active learning and transfer learning, so it can be combined with them for further improved performance. We have used ARIMLE to combine base classifiers constructed by transfer learning in this paper, and will integrate them with active learning in the future.

% Generated by IEEEtranS.bst, version: 1.14 (2015/08/26)

%\bibliographystyle{IEEETranS}\bibliography{drwubib}
\end{document}